\documentclass[10pt,twocolumn,letterpaper]{article}

\usepackage{wacv}              

\usepackage{float}
\usepackage{booktabs}
\usepackage{dblfloatfix}
\usepackage{multirow}
\usepackage{placeins}
\usepackage{soul}
\sethlcolor{yellow}

\definecolor{wacvblue}{rgb}{0.21,0.49,0.74}
\usepackage[breaklinks,colorlinks,allcolors=wacvblue]{hyperref}
\usepackage{graphicx}
\usepackage{amsmath}

\def\wacvPaperID{939} 
\def\confName{WACV}
\def\confYear{2027}

\title{MedicalDPO: Multi-Objective Direct Preference Optimization for CT to PET Translation}

\author{
Huiting Zhang$^{1}$\thanks{Equal contribution.} \quad
Yingxin Cai$^{1*}$ \quad
Siyuan Wang$^{1*}$ \\
Xi Yang$^{1}$ \thanks{Corresponding author.}\quad
Yuyao Yan$^{1}$ \quad
Rui Zhang$^{1}$ \quad
Kaizhu Huang$^{2}$\\
$^{1}$ Xi'an Jiaotong-Liverpool University, Suzhou, China\quad
$^{2}$ Duke Kunshan University, Kunshan, China\\ 
{\ttfamily\small
\{huiting.zhang25, yingxin.cai24, siyuan.wang21\}@student.xjtlu.edu.cn}
\\ {\ttfamily\small\{xi.yang01, yuyao.yan, rui.zhang02\}@xjtlu.edu.cn
\quad  kaizhu.huang@dukekunshan.edu.cn
}
}

\begin{document}
\maketitle
\footnotetext{This work has been published in WACV. The official published version is available on IEEE Xplore.}
\begin{abstract}
CT-to-PET synthesis estimates functional PET images from anatomical CT scans for auxiliary analysis, representation pretraining, and missing-modality completion. Brownian Bridge diffusion models can produce multiple PET candidates of different quality under the same CT condition, exposing an intra-condition quality hierarchy. Candidate quality spans complementary spatial, structural, and intensity dimensions, while DPO-only fine-tuning may compromise absolute synthesis fidelity. We propose \textbf{MedicalDPO} which converts this hierarchy into preference supervision automatically constructed from paired CT-PET training data. Multiple candidates are ranked across the three fidelity dimensions, and strict cross-metric verification retains preferred-rejected pairs with consistent quality separation. 
Relative denoising-error preference optimization is jointly trained with the original Brownian Bridge objective to anchor the pretrained CT-to-PET mapping, while condition-balanced multi-pair rotation distributes preference supervision across CT conditions. 
Experiments on BBDM and CPDM show improvements across LPIPS, MAE, SSIM, and PSNR. Further analyses demonstrate metric complementarity and clear pair separation, while ablations show that the original training objective prevents the substantial degradation caused by DPO-only fine-tuning.

\end{abstract}


\section{Introduction}
\label{sec:introduction}





CT-to-PET synthesis aims to estimate functional positron emission tomography (PET) images from anatomical computed tomography (CT) scans~\cite{cpdm}. This cross-modal translation can support auxiliary image analysis, representation pretraining, and missing-modality completion in multimodal medical imaging systems~\cite{missing,synthetic}. Its practical value depends on whether the synthesized PET images faithfully recover the spatial distribution, structural appearance, and intensity of metabolic uptake~\cite{synthetic}. However, CT primarily describes anatomical structure, whereas PET reflects functional activity that is not fully observable from CT~\cite{cpdm}. This cross-modal information gap makes reliable CT-to-PET synthesis fundamentally challenging.


\begin{figure}[!t]
    \centering
    \includegraphics[width=\linewidth]{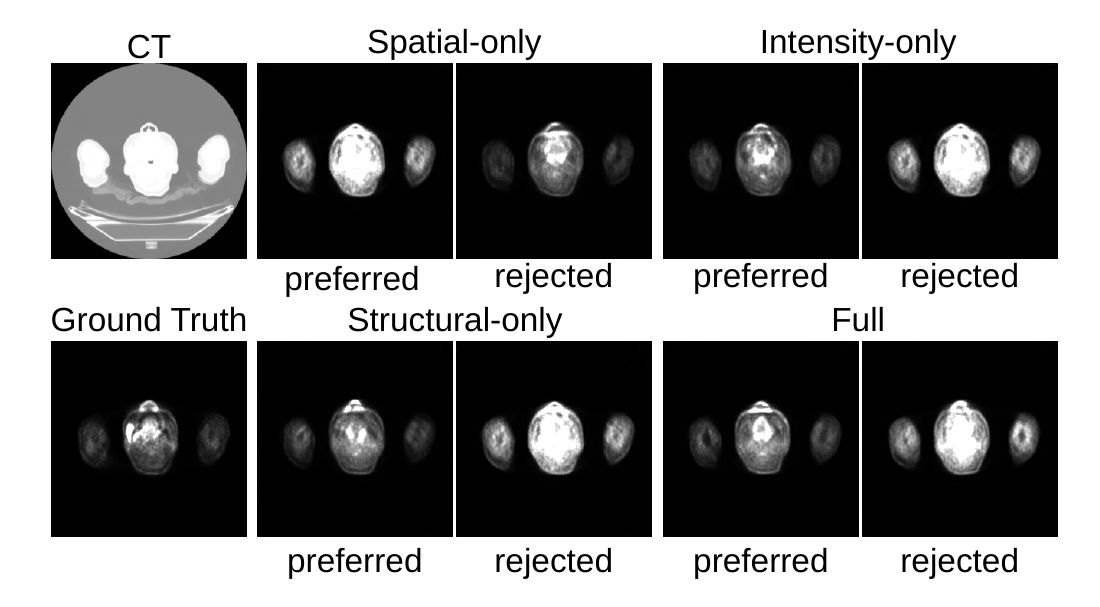}
    \caption{
Different fidelity criteria select preferred and rejected PET
candidates from $20$ CPDM-generated candidates for the same CT condition.
The full strategy combines spatial, structural, and intensity fidelity for
more reliable preference construction.
}
    \label{fig:intro_preference}
\end{figure}

Diffusion-based image translation models have shown strong potential for CT-to-PET synthesis. Brownian Bridge Diffusion Models
(BBDMs) formulate image translation as a stochastic bridge directly connecting the source and target domains~\cite{bbdm}, while the Conditional PET Diffusion Model (CPDM) incorporates CT-specific attention and attenuation guidance for PET generation~\cite{cpdm}. Despite their promising performance, stochastic generation can produce PET candidates of substantially different quality under the same CT condition~\cite{bbdm,Deterministic}.
Some candidates accurately recover high-uptake regions and global PET structure, while others exhibit localization errors, missing functional patterns, or intensity deviations.  Standard diffusion training optimizes each target sample independently and leaves these relative quality differences among generated candidates unused~\cite{bbdm,cpdm}.
This observation leads to two key insights. First, stochastic generation exposes an \emph{intra-condition quality hierarchy}. Multiple PET samples generated from the same CT provide automatically available preference supervision. A model can therefore learn not only from the ground-truth PET image, but also from the relative quality ordering of its own generated candidates. Second, PET candidate quality forms a \emph{vector, not a scalar}. Spatial fidelity reflects the localization of uptake regions, intensity fidelity measures the accuracy of tracer uptake values, and structural fidelity characterizes overall PET appearance~\cite{synthetic,generating}. These dimensions describe complementary properties and may assign different rankings to the same candidate set, as shown in Fig.~\ref{fig:intro_preference}. These complementary rankings motivate a joint representation of spatial, intensity, and structural fidelity for preference construction.


Learning from these candidate-level preferences introduces three key challenges. 
First, medical preference construction must integrate complementary fidelity dimensions while excluding ambiguous pairs. A candidate favored by one metric may still contain substantial errors in another dimension, making a single-metric preference potentially incomplete~\cite{generating,five_pitfalls}.  Second, Direct
Preference Optimization (DPO)~\cite{Original_dpo} optimizes relative preference differences but does not explicitly preserve absolute synthesis fidelity.  DPO-only fine-tuning may therefore shift the model away from its pretrained CT-to-PET mapping. 
Third, a single CT condition may produce multiple valid preference pairs. Treating all pairs as independent training samples overweights conditions with more valid combinations, motivating condition-balanced pair selection during optimization.

To address these challenges, we propose \textbf{MedicalDPO}, a 
preference-alignment framework for Brownian Bridge-based CT-to-PET synthesis.
Given a CT condition, MedicalDPO evaluates multiple generated PET candidates along spatial, structural, and intensity fidelity dimensions. Multi-dimensional ranking identifies high- and low-quality candidate pools, while strict cross-metric verification retains pairs with consistent quality separation across all fidelity measurements. Following Diffusion-DPO~\cite{diffusion_dpo}, the resulting preference pairs are used to optimize relative policy-reference denoising errors under the Brownian Bridge process. The preference objective is jointly optimized with the original Brownian Bridge training objective to preserve absolute synthesis fidelity. For CT conditions associated with multiple valid pairs, MedicalDPO rotates one pair per condition across training epochs, exposing the model to diverse preference supervision while maintaining condition-level balance.

MedicalDPO is evaluated on two Brownian Bridge-based CT-to-PET models, BBDM~\cite{bbdm} and the domain-guided CPDM~\cite{cpdm}. Quantitative and qualitative results demonstrate improved synthesis quality across both backbones. The complementarity and quality separation of the constructed preference pairs are further examined, while ablation studies assess the effects of the original training objective and multi-pair rotation.


Our main contributions are summarized as follows:
\begin{itemize}
    \item We formulate stochastic CT-to-PET refinement as intra-condition preference learning. Multiple PET candidates generated from the same CT expose a quality hierarchy that provides automatically available supervision, with candidate quality represented across complementary spatial, structural, and intensity dimensions.    
    
    \item We introduce a multi-objective preference construction strategy that combines multi-dimensional candidate ranking with strict cross-metric verification, retaining strict pairs with cross-metric-consistent quality separation.

    \item We develop a stable preference adaptation framework for Brownian Bridge models by combining relative denoising-error preference optimization with the original training objective and condition-balanced multi-pair rotation. Experiments on BBDM and CPDM demonstrate improvements across perceptual, intensity, structural, and reconstruction metrics.
\end{itemize}
\section{Related Work}
\label{sec:related_work}

\subsection{CT-to-PET Translation}
Image-to-image (I2I) translation learns mappings between visual domains,
with paired approaches such as Pix2Pix~\cite{pix2pix}.
GAN-based methods have also been explored for CT-to-PET
translation~\cite{GAN}, but may suffer from training instability and
mode collapse~\cite{gan_instability}, motivating diffusion-based
alternatives~\cite{ddpm}.
BBDM~\cite{bbdm} formulates I2I translation as a Brownian Bridge
directly connecting the source and target domains, while
CPDM~\cite{cpdm} adapts this framework to CT-to-PET synthesis using
domain-specific attention and attenuation guidance.
Despite strong performance, both are trained mainly with per-sample diffusion objectives and don’t explicitly use relative quality differences among stochastic PET candidates, motivating preference-based refinement.

\subsection{Preference Optimization for Diffusion Models}
Preference optimization provides a natural way to align generative
models with desired output characteristics. RLHF learns a reward model
from pairwise comparisons and optimizes it through reinforcement
learning~\cite{RLHF}, whereas DPO~\cite{Original_dpo} learns directly from
preferred and rejected samples without an explicit reward model.
For diffusion models, prior work has explored reinforcement-learning
reward maximization~\cite{ddpo,dpok} and preference-based
optimization~\cite{diffusion_dpo,smpo}. In particular,
Diffusion-DPO~\cite{diffusion_dpo} adapts DPO to diffusion models using
pairwise preference data. 
Existing diffusion preference methods mainly target general-purpose or
text-to-image generation, where preferences are commonly defined by
human judgments, visual appeal, or text-image alignment. Medical image
translation instead requires task-specific preferences that jointly
reflect spatial, intensity, and structural fidelity, motivating the
preference formulation developed in this work.
\begin{figure*}[!t]
    \centering
    \includegraphics[width=\textwidth]{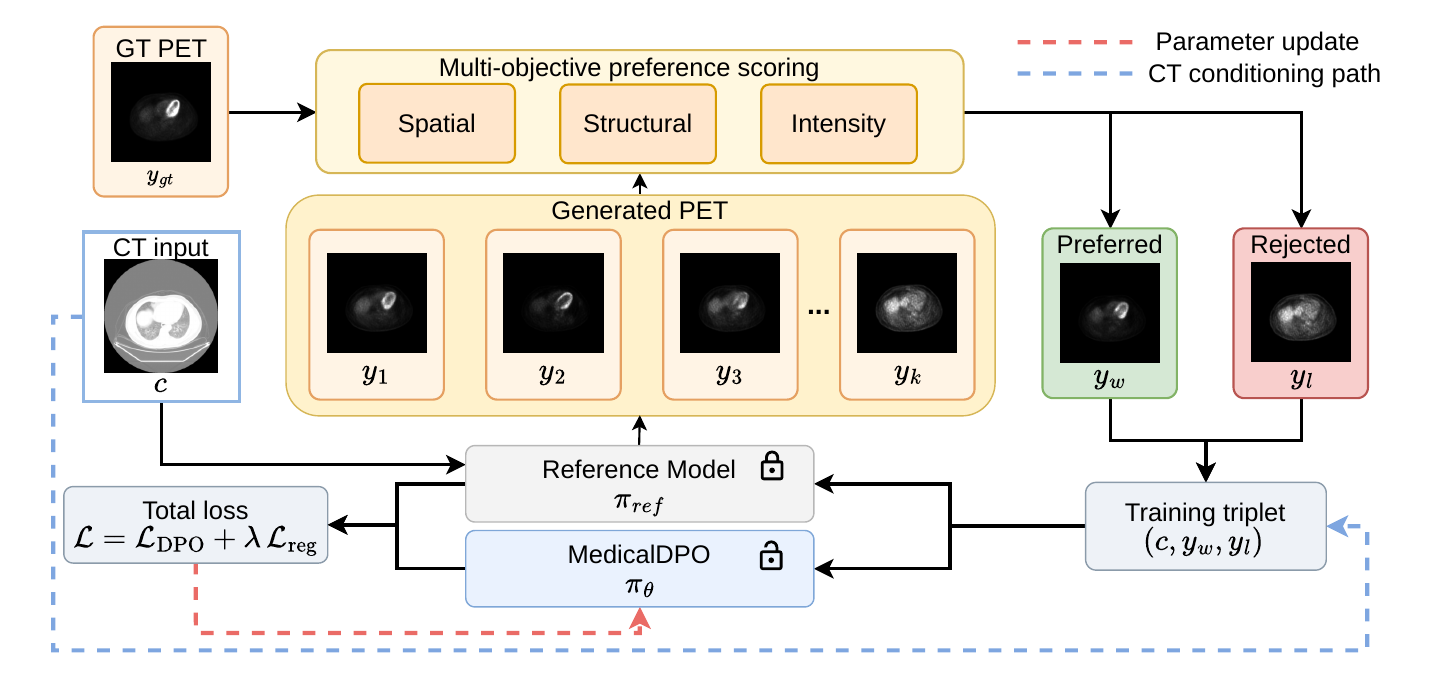}
   \caption{
Overview of the proposed MedicalDPO framework for CT-to-PET synthesis,
built on Brownian Bridge diffusion models~\cite{bbdm,cpdm}.
The figure visualizes the corresponding CT and PET images, while
$c$ denotes the latent CT conditioning representation used by the
Brownian Bridge model. Multiple generated PET candidates are evaluated
against the ground-truth PET using complementary spatial, structural,
and intensity criteria to construct reliable preference pairs.
The selected preferred and rejected samples are encoded as $y_w$ and
$y_l$, forming the latent preference triplet $(c,y_w,y_l)$ for optimizing
the trainable policy $\pi_\theta$ against the frozen reference model
$\pi_{\rm ref}$ with
$\mathcal{L}=\mathcal{L}_{\rm DPO}+\lambda\mathcal{L}_{\rm reg}$.
Open/closed locks denote trainable/frozen components.
}
    \label{fig:framework}
\end{figure*}
\section{Method}
\label{sec:method}

In this section, we present a preference-aligned framework for CT-to-PET synthesis that
constructs reliable multi-objective preference pairs and adapts preference
optimization to Brownian Bridge diffusion models~\cite{bbdm,cpdm}.

\subsection{Preliminaries}
\label{sec:preliminaries}
\noindent\textbf{Brownian Bridge Diffusion Models.}
\label{sec:brownian_bridge}
BBDM~\cite{bbdm} models image-to-image translation
through a Brownian Bridge diffusion process, while
CPDM~\cite{cpdm} extends this framework to CT-to-PET
synthesis by incorporating domain-specific conditioning. 
Following latent diffusion models~\cite{latent}, the diffusion process is conducted in a latent space rather than the original image space. Within this space, let $c$ and $y$ denote the latent representations of a given pair of CT and PET images, respectively, which are encoded via a pretrained VQGAN~\cite{VQGAN}.
During the forward process,
the intermediate latent state at timestep $t$
is defined as
\begin{equation}
x_t = (1-m_t)y + m_t c + \sqrt{\delta_t}\epsilon,
\label{eq:bb_forward}
\end{equation}
where $m_t=t/T$, $T$ denoted as the total denoising step. $\delta_t$ represents the Brownian Bridge
variance, and $\epsilon\sim\mathcal{N}(0,I)$ is Gaussian noise.


For preference optimization, we define the following mean squared prediction error under the Brownian Bridge forward process~\cite{bbdm}
\begin{equation}
\ell_{\mathrm{BB},\theta}
=
\frac{1}{d}
\left\|
m_t(c-y)+\sqrt{\delta_t}\epsilon
-\hat{\epsilon}_\theta(x_t,t,c)
\right\|_2^2,
\label{eq:cpdm_loss}
\end{equation}
where $d$ denotes the number of latent elements, and $\hat{\epsilon}_\theta$ is predicted by the denoising network.



\noindent\textbf{Direct Preference Optimization.}
DPO~\cite{Original_dpo} directly optimizes pairwise preferences without a
separate reward model. 
The original objective is
\begin{equation}
\mathcal{L}_{\mathrm{DPO}}^{\mathrm{orig}}
=
-\log \sigma
\left[
\beta
\left(
\log \frac{\pi_\theta(y_w|c)}{\pi_{\mathrm{ref}}(y_w|c)}
-
\log \frac{\pi_\theta(y_l|c)}{\pi_{\mathrm{ref}}(y_l|c)}
\right)
\right],
\label{eq:original_dpo}
\end{equation}
where $\pi_\theta$ and $\pi_{\mathrm{ref}}$ denote the policy and frozen
reference models, respectively, and $\beta$ controls the preference strength. $c$ represents the CT condition and \(y_w\)$y_w$ and $y_l$ denote the preferred and rejected PET candidates generated under the same condition.

\subsection{Multi-objective Preference Construction}
\label{sec:preference}
\noindent\textbf{Multi-dimensional Preference Ranking.}
PET candidates may differ in spatial localization, intensity, and structure,
making single-metric ranking potentially biased. We therefore construct
preferences across these three complementary fidelity dimensions, as
validated in Sec.~\ref{sec:preference_analysis}. For each CT condition, 20 PET candidates are generated by the corresponding frozen reference backbone using independent sampling noise.
Let $Y_s$ denote the $s$-th generated PET candidate in image space,
and let $Y_{\rm gt}$ denote the corresponding ground-truth PET image.
Preference pairs are constructed separately for BBDM and CPDM.

\noindent\textbf{Spatial Fidelity} measures high-uptake localization
using the Dice coefficient~\cite{Dice} between generated and
ground-truth PET masks
\begin{equation}
\operatorname{Dice}(Y_s,Y_{\rm gt})
=
\frac{
2|M(Y_s)\cap M(Y_{\rm gt})|
}{
|M(Y_s)|+|M(Y_{\rm gt})|
}
\end{equation}
where $M(\cdot)$ denotes the high-uptake mask, obtained by scaling PET
intensities to $[0,255]$ with a fixed maximum of $32767$ and thresholding
pixels above $100$.

\noindent\textbf{Intensity Fidelity} measures the intensity discrepancy
between generated PET images and ground truth. For preference construction, MAE and WMAE are computed on normalized PET intensities. We first employ mean absolute error (MAE) to quantify the global pixel-wise intensity error
\begin{equation}
\operatorname{MAE}(Y_s,Y_{\rm gt})
=
\frac{1}{N}
\sum_{j=1}^{N}
|Y_{s,j}-Y_{{\rm gt},j}|
\end{equation}
where $N$ is the number of pixels.
To emphasize errors in regions highlighted by the attention map~\cite{cpdm},
we further use attention-weighted MAE (WMAE)
\begin{equation}
\operatorname{WMAE}(Y_s,Y_{\rm gt})
=
\frac{
\sum_{j=1}^{N} A_j |Y_{s,j}-Y_{{\rm gt},j}|
}{
\sum_{j=1}^{N} A_j
}
\end{equation}
where $A_j \ge 0$ denotes the attention weight at pixel $j$. Compared with MAE, WMAE assigns larger importance to intensity
errors in attention-focused regions, providing a more
region-aware measurement of PET intensity fidelity.
The attention maps are precomputed and kept fixed during MedicalDPO
fine-tuning, with the same map used for all candidates of a given CT
condition. Further implementation details are provided in
Sec.~B.3 of the supplementary material.

\noindent\textbf{Structural Fidelity} measures consistency between
generated and ground-truth PET using SSIM~\cite{ssim}, which evaluates
local structure, contrast, and luminance
\begin{equation}
\mathrm{SSIM}(Y_s,Y_{gt})
=
\frac{
(2\mu_{Y_s}\mu_{gt}+C_1)
(2\sigma_{Y_s,Y_{gt}}+C_2)
}
{
(\mu_{Y_s}^{2}+\mu_{gt}^{2}+C_1)
(\sigma_{Y_s}^{2}+\sigma_{gt}^{2}+C_2)
}
\label{eq:ssim}
\end{equation}
where $\mu$, $\sigma^2$, and $\sigma_{y_i y_{gt}}$ denote the local means,
variances, and covariance, respectively, and $C_1,C_2$ are stability constants.
Higher SSIM indicates better structural fidelity.

\noindent\textbf{Rank Aggregation.}
We rank candidates independently along the three fidelity dimensions. Dice~\cite{Dice} and SSIM~\cite{ssim} are ranked in descending order,
whereas MAE and WMAE are ranked in ascending order, with lower ranks
indicating better fidelity. Since MAE and WMAE both measure intensity
fidelity, we average their ranks to avoid overweighting intensity
\begin{equation}
r_{\rm int}=\frac{r_{\rm MAE}+r_{\rm WMAE}}{2}.
\end{equation}

The final aggregate rank is defined by equally combining the spatial,
intensity, and structural ranks
\begin{equation}
S_{\mathrm{rank}}
=
\frac{1}{3}
\left(
r_{\mathrm{spatial}}
+
r_{\mathrm{int}}
+
r_{\mathrm{struct}}
\right),
\label{eq:overall_rank}
\end{equation}
where lower $S_{\rm rank}$ indicates better overall quality.

\noindent\textbf{Winner-Loser Pair Selection.}
To construct reliable preference pairs, we partition the ranked candidates according to both aggregate quality and cross-objective consistency. Candidates satisfying the high- and low-quality criteria form the good and bad pools, respectively, while the remaining candidates are regarded as ambiguous and excluded from preference construction.

Formally, the good and bad pools satisfy
\begin{equation}
\begin{aligned}
\mathcal{G}:&\quad S_{\mathrm{rank}}\leq\tau_g,
\quad n_{\mathrm{top}}\geq2,\\
\mathcal{B}:&\quad S_{\mathrm{rank}}\geq\tau_b,
\quad n_{\mathrm{bottom}}\geq2,
\end{aligned}
\label{eq:quality_pool}
\end{equation}
Here, $\tau_g=5$ and $\tau_b=16$. Let $\mathcal{F}=\{\mathrm{spatial},\mathrm{int},\mathrm{struct}\}$,
and let $r_f(Y_s)$ denote the rank of candidate $Y_s$
under fidelity dimension $f$.
We compute
\begin{equation}
\begin{aligned}
n_{\rm top}(Y_s)
&=
\left|\{f\in\mathcal{F}:r_f(Y_s)\leq5\}\right|,\\
n_{\rm bottom}(Y_s)
&=
\left|\{f\in\mathcal{F}:r_f(Y_s)\geq16\}\right|.
\end{aligned}
\label{eq:dimension_count}
\end{equation}
Thus, good/bad candidates must fall within the top-/bottom-5 aggregate
ranks and be supported by at least two of the three fidelity dimensions.

For multi-pair construction, we select up to two best-ranked candidates
from the good pool as preferred candidates and up to two worst-ranked
candidates from the bad pool as rejected candidates, yielding at most
four cross-pool combinations for each CT condition.
We denote a selected preferred candidate by $Y_w$ and a selected
rejected candidate by $Y_l$. To suppress ambiguous supervision, only
combinations in which the preferred sample consistently outperforms the
rejected sample across all four fidelity measurements are retained
\begin{equation}
\begin{aligned}
\mathrm{Dice}(Y_w) &> \mathrm{Dice}(Y_l), &
\mathrm{SSIM}(Y_w) &> \mathrm{SSIM}(Y_l),\\
\mathrm{MAE}(Y_w) &< \mathrm{MAE}(Y_l), &
\mathrm{WMAE}(Y_w) &< \mathrm{WMAE}(Y_l). \nonumber
\end{aligned}
\label{eq:strict_pair}
\end{equation}

Only pairs satisfying all four constraints are retained as strict
preferences. Their corresponding latent representations are used for
DPO optimization, as described in Sec.~\ref{sec:dpo}.
The reliability of the resulting preference pairs is further evaluated
quantitatively and qualitatively in
Sec.~\ref{sec:preference_analysis}.

\subsection{Direct Preference Optimization for Brownian Bridge Models}
\label{sec:dpo}

The preferred and rejected PET images ($Y_w$, $Y_l$) and the corresponding CT image $C$ are encoded into latent representations $y_w$, $y_l$, and $c$. The latent triplet $(c,y_w,y_l)$ is then used for preference optimization.
Applying the likelihood-based DPO objective in
Eq.~\ref{eq:original_dpo} to diffusion models is non-trivial
because exact likelihoods require marginalizing the diffusion
trajectory~\cite{diffusion_dpo}. Inspired by
Diffusion-DPO~\cite{diffusion_dpo}, we use a preference surrogate based on policy–reference prediction-error differences under the Brownian Bridge forward process~\cite{bbdm,cpdm}.
Preferred and rejected samples are evaluated at the same timestep $t$ with shared Gaussian noise $\epsilon$, reducing variability in their relative denoising errors.
The policy-reference denoising error difference is defined as
\begin{equation}
\Delta_{\theta}^{\mathrm{BB}}(y,c;t,\epsilon)
=
\ell_{\mathrm{BB},\theta}(y,c;t,\epsilon)
-
\ell_{\mathrm{BB},\mathrm{ref}}(y,c;t,\epsilon),
\label{eq:bb_preference_margin}
\end{equation}
where $\ell_{\mathrm{BB},\theta}$
denotes the denoising error defined in Eq. \ref{eq:cpdm_loss} evaluated using the trainable policy model, and 
$\ell_{\mathrm{BB},\mathrm{ref}}$ is computed by the frozen reference
model. 

Based on the policy-reference denoising error gap, we formulate the MedicalDPO
objective as
\begin{equation}
\mathcal{L}_{\rm DPO}
=
-\mathbb{E}_{(c,y_w,y_l),t,\epsilon}
\left[
\log \sigma
\left(
-\frac{\beta}{2}
\left(
\Delta_w^{\rm BB}-\Delta_l^{\rm BB}
\right)
\right)
\right],
\end{equation}
where
$\Delta_w^{\mathrm{BB}}=
\Delta_\theta^{\mathrm{BB}}(y_w,c;t,\epsilon)$
and
$\Delta_l^{\mathrm{BB}}=
\Delta_\theta^{\mathrm{BB}}(y_l,c;t,\epsilon)$
are policy-reference denoising error differences for the preferred and rejected
samples, respectively.

\subsection{Training Strategy}
\label{sec:training_strategy}
\label{sec:training}

\noindent\textbf{Preference Pair Scheduling.}
We consider two strategies for using strict preference pairs.
\textit{Single} fixes one retained strict pair for each CT condition
throughout fine-tuning. In contrast, when multiple strict pairs are
available, \textit{Rotation} selects one pair per CT per epoch and
cyclically rotates through its retained set
$\mathcal{P}_i=\{(y^w_{i,k},y^l_{i,k})\}_{k=1}^{K_i}$,
where $i$ indexes CT conditions, $k$ indexes the retained preference
pairs for condition $i$, and $K_i$ denotes the number of retained
pairs for that condition, with $1\le K_i\le4$. This avoids over-representing CTs with more valid pairs.
During validation, all $K_i$ pairs are evaluated with weight $1/K_i$,
ensuring equal CT contribution.

\noindent\textbf{Regularized Preference Optimization.}
To preserve pretrained generation behavior during preference fine-tuning, we combine DPO with the original training objective of the corresponding Brownian Bridge backbone~\cite{bbdm,cpdm}
\begin{equation}
\mathcal{L}
=
\mathcal{L}_{\mathrm{DPO}}
+
\lambda \mathcal{L}_{\mathrm{reg}},
\end{equation}
where $\mathcal{L}_{\rm reg}$ is the original backbone loss and $\lambda$
controls the trade-off between preference alignment and pretrained-model
preservation.

\section{Experiments}
\label{sec:experiments}








\subsection{Experimental Setup}
\label{sec:Setup}
We evaluate MedicalDPO for CT-to-PET translation on two diffusion-based backbones, BBDM~\cite{bbdm} and CPDM~\cite{cpdm}, with Pix2Pix~\cite{pix2pix} included as a GAN-based baseline.

\noindent\textbf{Training Dataset.}
We construct our dataset from two publicly available whole-body PET/CT
datasets, FDG-PET-CT-Lesions~\cite{fdg_pet_ct} and
ViMed-PET~\cite{vimed_pet}. After preprocessing, the training split contains
111,630 paired CT and attenuation-corrected PET slices.
Pix2Pix~\cite{pix2pix}, BBDM~\cite{bbdm}, and CPDM~\cite{cpdm} are trained
on this full training split.
For preference-based fine-tuning, we sample 15,000 CT--PET conditions
exclusively from the training split and construct backbone-specific preference
sets independently. For CPDM, the final training preference set contains
48,545 pairs from 12,704 CT conditions. For BBDM, strict 4/4 filtering retains 14,094 CT conditions, which are split patient-wise into 12,685
training and 1,409 validation conditions with no patient overlap.

\noindent\textbf{Validation and Test Data.}
A separate validation set of 1,500 paired CT--PET slices is used to select base-model checkpoints. For BBDM and CPDM, the best checkpoints on this validation set are used as the pretrained models for subsequent MedicalDPO fine-tuning. All methods are evaluated on a test set of 246 paired CT--PET slices from the data released with CPDM~\cite{cpdm}.
The test set is used only for final evaluation and is excluded from preference construction.
The base-model training, validation, and test splits are patient-disjoint.
For CPDM and BBDM, retained preference conditions are also split patient-wise into training and validation sets with no patient overlap.

\noindent\textbf{Baselines.}
Pix2Pix~\cite{pix2pix} is used as a GAN-based baseline, while
BBDM~\cite{bbdm} and CPDM~\cite{cpdm} serve as the pretrained
Brownian Bridge backbones for MedicalDPO.

\noindent\textbf{Metrics.}
We evaluate synthesis quality using LPIPS~\cite{LPIPs}, MAE~\cite{Synthetic_PET},
SSIM~\cite{ssim}, and PSNR~\cite{medgan,uncertainty_guidedgan}, which
measure perceptual similarity, pixel-wise intensity error, structural
similarity, and reconstruction fidelity, respectively. Lower LPIPS/MAE and
higher SSIM/PSNR indicate better performance. Evaluation MAE uses the PET scale, whereas preference MAE/WMAE use normalized
intensities.

\noindent\textbf{Implementation Details.}
All images are resized to $256\times256$. For preference scoring, attention maps are generated using the CPDM
attention-map pipeline~\cite{cpdm}, using a
U-Net~\cite{unet} with a pretrained ResNet-34~\cite{resnet} backbone.
The same attention maps are used to construct preferences for both 
CPDM- and BBDM-generated candidates. MedicalDPO is initialized from pretrained BBDM~\cite{bbdm} and CPDM~\cite{cpdm}, then fine-tuned with AdamW~\cite{adamw} (lr $1\times10^{-7}$, $\beta=100$, $\lambda=11$, batch size 8). Unless noted, we report the epoch-4 checkpoint for the 15K preference setting. Preference candidates are generated using 50 sampling steps, while final evaluation uses 200 steps for both BBDM and CPDM, averaged over three seeds (42, 1234, 2025). Base BBDM/CPDM are trained on NVIDIA RTX 4090 GPUs, while Pix2Pix~\cite{pix2pix} and MedicalDPO use RTX 2080 Ti GPUs. The 20 preference candidates are generated offline before fine-tuning  as a one-time cost. No extra inference-time modules are required, and the pretrained inference architecture is unchanged.

\subsection{Analysis of Preference Pair Construction}
\label{sec:preference_analysis}





\noindent\textbf{Quantitative Analysis.}
The proposed preference construction identifies higher-quality PET candidates by combining complementary fidelity measurements with cross-metric consistency. We first examine whether the selected metrics provide distinct ranking information for candidates from the same CT condition. As reported in Tab.~\ref{tab:metric_complementarity}, Dice shows low agreement with the other metrics, with best-candidate agreement ranging from 15.9\% to 21.6\%, indicating that spatial fidelity captures information not fully reflected by structural or intensity measurements. In contrast, MAE and WMAE ~\cite{Synthetic_PET} show the strongest agreement, with a mean Spearman correlation of 0.880 and best-candidate agreement of 59.8\%, supporting their aggregation into a single intensity-fidelity dimension and the use of complementary objectives for preference ranking.

\begin{table}[t]
\centering
\caption{
Complementarity of the selected preference metrics.
$\rho$ denotes the mean within-sample Spearman correlation over valid cases,
and Agree. denotes the percentage of valid CT cases for which two metrics
select the same best candidate.
}
\label{tab:metric_complementarity}
\small
\setlength{\tabcolsep}{5pt}
\begin{tabular}{lcc}
\toprule
Metric Pair & Mean $\rho$ & Agree. (\%) \\
\midrule
Dice--SSIM & 0.335 & 15.9 \\
Dice--MAE & 0.400 & 21.6 \\
Dice--WMAE & 0.383 & 20.1 \\
SSIM--MAE & 0.764 & 44.0 \\
SSIM--WMAE & 0.657 & 32.3 \\
MAE--WMAE & \textbf{0.880} & \textbf{59.8} \\
\bottomrule
\end{tabular}
\end{table}

We further assess the quality separation between preferred and rejected candidates. As reported in Tab.~\ref{tab:preference_separation}, preferred candidates achieve higher Dice (0.512 vs.\ 0.217) and SSIM (0.915 vs.\ 0.874), and lower MAE (0.0095 vs.\ 0.0176) and WMAE (0.0305 vs.\ 0.0595). These margins demonstrate clear separation across all three fidelity dimensions. Since retained pairs satisfy the strict 4/4 criterion by construction, these statistics mainly quantify the magnitude
of the quality gap between preferred and rejected samples.
Additional preference-quality analysis of BBDM-generated candidates is in Sec.~C.2 of the suppl. material.

\begin{table}[t]
\centering
\caption{
Sample-balanced quality separation between preferred and rejected PET
candidates over 48,545 strict preference pairs from 12,704 CT conditions.
Pair weights ensure equal contribution from each CT. MAE and WMAE are computed on normalized PET intensities.
$\Delta$ denotes Preferred$-$Rejected; negative values for MAE and WMAE
indicate reduced errors. All retained pairs satisfy the strict 4/4 criterion.
}
\label{tab:preference_separation}
\small
\setlength{\tabcolsep}{3.5pt}
\begin{tabular}{lccc}
\toprule
Metric & Preferred $y_w$ & Rejected $y_l$ & $\Delta$ \\
\midrule
Dice $\uparrow$
& \textbf{0.512$\pm$0.238}
& 0.217$\pm$0.234
& +0.295 \\

SSIM $\uparrow$
& \textbf{0.915$\pm$0.040}
& 0.874$\pm$0.110
& +0.041 \\

MAE $\downarrow$
& \textbf{0.0095$\pm$0.0049}
& 0.0176$\pm$0.0103
& -0.0081 \\

WMAE $\downarrow$
& \textbf{0.0305$\pm$0.0252}
& 0.0595$\pm$0.0543
& -0.0290 \\
\bottomrule
\end{tabular}
\end{table}

\noindent\textbf{Qualitative Analysis.}
The example in Fig.~\ref{fig:intro_preference} further
illustrates the behavior of the proposed preference construction.
Preferences obtained from individual fidelity criteria may favor different
PET candidates, whereas the full multi-objective strategy selects a preferred
candidate with more consistent spatial localization, structural appearance,
and intensity distribution with respect to the ground truth.
Additional single-objective qualitative comparisons are 
in Sec.~C.2 of the suppl. material.

\subsection{Experimental Results}
\label{sec:experimental_results}

\subsubsection{Qualitative Results}
\label{sec:qualitative_results}

%
\begin{figure*}[t]
    \centering
    \includegraphics[width=0.98\textwidth]{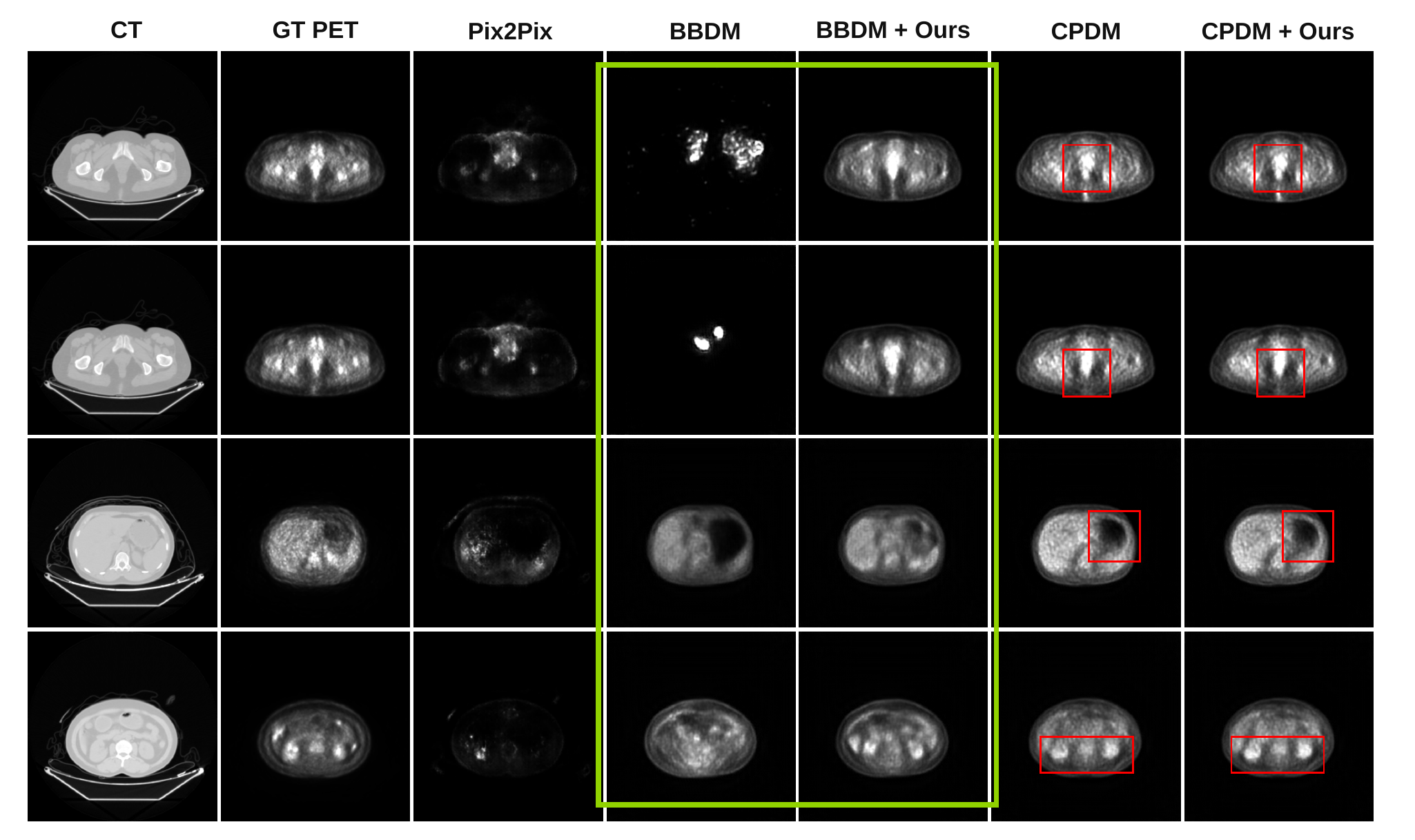}
    \caption{
    Qualitative comparison of CT-to-PET synthesis methods. The green box highlights the overall improvement of BBDM~\cite{bbdm} after applying MedicalDPO, while the red boxes emphasize local improvements over CPDM~\cite{cpdm}, better matching the ground-truth PET.
    }
    \label{fig:qualitative_results}
\end{figure*}

As illustrated in \cref{fig:qualitative_results}, MedicalDPO improves both
Brownian-Bridge backbones, although the visual gains differ with the
strength of the pretrained model. For BBDM~\cite{bbdm}, the pretrained model produces incomplete or distorted PET patterns, whereas MedicalDPO recovers coherent structures and tracer distributions,
as highlighted by the green box. CPDM~\cite{cpdm}, aided by domain-specific attention and attenuation guidance, preserves stronger global structure;  nevertheless, MedicalDPO further improves
local uptake patterns and structural details, as highlighted by the red
boxes. The results show improved global fidelity for BBDM and finer local consistency for CPDM.

\begin{table}[t]
\centering
\caption{
Quantitative comparison of CT-to-PET synthesis methods. All MedicalDPO variants use epoch-4 checkpoints. Bold denotes the best result for each backbone and Diff. denotes the relative performance change from the corresponding base model. $\downarrow/\uparrow$ indicates lower/higher is better.
}
\label{tab:main_results}

\small
\setlength{\tabcolsep}{4.0pt}
\renewcommand{\arraystretch}{1.08}

\begin{tabular}{lcccc}
\toprule
Method 
& LPIPS $\downarrow$ 
& MAE $\downarrow$ 
& SSIM $\uparrow$ 
& PSNR $\uparrow$ \\
\midrule

Pix2Pix~\cite{pix2pix}
& 0.1465 
& 849.51 
& 0.8274 
& 22.96 \\

\midrule

BBDM~\cite{bbdm}
& 0.1131
& 711.38
& 0.8405
& 24.34 \\

+Ours (Single)
& 0.1006
& 626.42
& 0.8591
& 25.02 \\

+Ours (Rotation)
& \textbf{0.0999}
& \textbf{617.31}
& \textbf{0.8594}
& \textbf{25.14} \\

\midrule

Diff. (Single)
& $-11.05\%$
& $-11.94\%$
& $+2.21\%$
& $+2.79\%$ \\

Diff. (Rotation)
& $-11.67\%$
& $-13.22\%$
& $+2.25\%$
& $+3.29\%$ \\

\midrule

CPDM~\cite{cpdm}
& 0.0976
& 751.05
& 0.8499
& 24.27 \\

+Ours (Single)
& \textbf{0.0967}
& \textbf{711.74}
& \textbf{0.8548}
& \textbf{24.42} \\

+Ours (Rotation)
& 0.0969
& 727.03
& 0.8537
& 24.30 \\

\midrule

Diff. (Single)
& $-0.92\%$
& $-5.23\%$
& $+0.58\%$
& $+0.62\%$ \\

Diff. (Rotation)
& $-0.72\%$
& $-3.20\%$
& $+0.45\%$
& $+0.12\%$ \\

\bottomrule
\end{tabular}
\end{table}

\begin{table*}[!t]
\setlength{\tabcolsep}{5.0pt}
\centering
\resizebox{0.92\textwidth}{!}{
\begin{tabular}{l||cccc||cccc}
\hline

Method
& \multicolumn{4}{c||}{BBDM~\cite{bbdm}}
& \multicolumn{4}{c}{CPDM~\cite{cpdm}}
\\

\cline{2-9}

& LPIPS $\downarrow$
& MAE $\downarrow$
& SSIM $\uparrow$
& PSNR $\uparrow$
& LPIPS $\downarrow$
& MAE $\downarrow$
& SSIM $\uparrow$
& PSNR $\uparrow$
\\
\hline

Base
& 0.1131 & 711.38 & 0.8405 & 24.34
& 0.0976 & 751.05 & 0.8499 & 24.27 \\

+ DPO 
& 0.1911 & 2143.13 & 0.5079 & 17.19
& 0.1392 & 1497.95 & 0.7928 & 19.36 \\

+ Only GT 
& 0.1111 & 701.33 & 0.8422 & 24.42
& 0.0975 & 755.23 & 0.8500 & 24.14 \\

\hline

+ DPO + Guard 
& 0.1128 & 703.91 & 0.8529 & 24.48
& 0.1193 & 1263.73 & 0.8223 & 20.51 \\

+ DPO + GT + Drift 
& 0.1000 & 622.71 & 0.8590 & 25.09
& 0.1133 & 1149.17 & 0.8301 & 21.12 \\

+ Winner-to-GT 
& 0.3532 & 1047.85 & 0.7011 & 21.81
& 0.2819 & 846.99 & 0.7758 & 22.99 \\

+ Winner/Loser-to-GT (A)
& 0.3463 & 1080.96 & 0.7146 & 21.38
& 0.2949 & 857.68 & 0.7738 & 22.94 \\

+ Winner/Loser-to-GT (B)
& 0.3557 & 1065.68 & 0.6998 & 21.67
& 0.2841 & 849.68 & 0.7765 & 22.97 \\

\hline

+ Ours (Single)
& 0.1006 & 626.42 & 0.8591 & 25.02
& \textbf{0.0967}
& \textbf{711.74}
& \textbf{0.8548}
& \textbf{24.42} \\

+ Ours (Rotation)
& \textbf{0.0999}
& \textbf{617.31}
& \textbf{0.8594}
& \textbf{25.14}
& 0.0969 & 727.03 & 0.8537 & 24.30 \\

\hline
\end{tabular}
}

\vspace{-2mm}

\caption{
Ablation of different preference fine-tuning strategies on BBDM and CPDM. All variants use epoch-4 checkpoints. Bold denotes the best result. $\downarrow/\uparrow$ indicates lower/higher is better.
}
\label{tab:ablation}

\end{table*}

\subsubsection{Quantitative Results}
\label{sec:quantitative_results}

Tab.~\ref{tab:main_results}  reports the quantitative comparisons on 
test set. MedicalDPO consistently improves both
BBDM and CPDM across all four metrics.
Multi-pair Rotation performs best on BBDM, reducing LPIPS and MAE by
11.67\% and 13.22\%, respectively, while Single performs best on CPDM,
with a 5.23\% reduction in MAE and consistent gains in the other metrics.
Multi-pair rotation performs best on BBDM. On CPDM, both strategies achieve similar results, with single-pair training performing slightly better.
Our approach improves both models across LPIPS~\cite{LPIPs} and SSIM~\cite{ssim} for perceptual and structural similarity, and MAE~\cite{Synthetic_PET} and PSNR~\cite{medgan,uncertainty_guidedgan} for reconstruction accuracy. More quantitative results are in the suppl. material.
\subsection{Ablation Study}
\label{sec:ablation}
\noindent\textbf{Optimization Strategy Ablation.}
We compare different preference fine-tuning strategies on BBDM
and CPDM, as shown in Tab.~\ref{tab:ablation}.
1) +DPO adapts Diffusion-DPO~\cite{diffusion_dpo} to the Brownian-Bridg formulation.
It uses the same preference pairs as MedicalDPO but removes the
proposed backbone-preserving regularization.
It substantially degrades both backbones, indicating that preference
optimization alone may disrupt the pretrained CT-to-PET mapping.
2) Only GT further fine-tunes the backbone using only ground-truth
supervision and remains close to the base models, showing that MedicalDPO's
gains are not simply due to additional training.
3) DPO + Guard constrains the preferred sample against degradation relative
to the reference model, while DPO + GT + Drift combines ground-truth
supervision with a constraint that limits deviation from the reference model.
These variants partially mitigate the degradation on BBDM but fail to
consistently improve CPDM, suggesting limited cross-backbone robustness.
4) Winner-to-GT adds an $\ell_1$ winner-to-GT loss with
$\lambda_{\rm gt}=11$, while Winner/Loser-to-GT (A) and (B) use
$(\lambda_{\rm gt},\lambda_{\rm losegt})=(11,11)$ and $(11,1.1)$,
respectively. These variants directly constrain preference samples toward
the ground truth but still degrade performance, indicating that GT proximity
alone is insufficient for effective alignment.
In contrast, MedicalDPO improves all four metrics on both backbones,
demonstrating more robust cross-backbone preference adaptation. Detailed formulations are provided in Sec.~A.3 of the supple. material.


\begin{figure}[t]
    \centering
    \includegraphics[width=\linewidth]
    {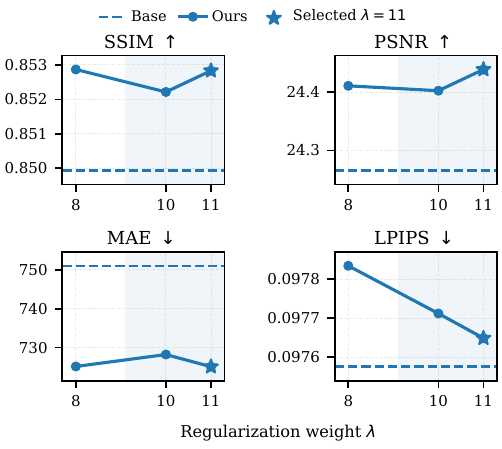}
    \caption{
Sensitivity to the regularization weight $\lambda$ on the CPDM backbone.
We evaluate $\lambda\in\{8,10,11\}$ under the same 40-epoch budget on the 2.5K preference subset used for hyperparameter selection.
The dashed line denotes Base CPDM, the shaded region the geometry-supported interval, and the star the selected $\lambda=11$.
$\lambda=11$ yields the best PSNR, MAE, and LPIPS, and is nearly best in SSIM.
}
    \label{fig:lambda_sensitivity}
\end{figure}

\noindent\textbf{Sensitivity to $\lambda$.}
We evaluate the regularization weight $\lambda$ on the
CPDM backbone, as shown in Fig.~\ref{fig:lambda_sensitivity}.
This analysis uses a 2.5K preference subset for hyperparameter selection,
independently from the final 15K experiments. All
$\lambda\in\{8,10,11\}$ settings are compared under the same 40-epoch
training budget; this epoch-40 setting is used only for hyperparameter
analysis, whereas the main 15K experiments report epoch-4 checkpoints
under the final training protocol. Among the evaluated values,
$\lambda=11$ achieves the best PSNR~\cite{medgan,uncertainty_guidedgan},
MAE~\cite{Synthetic_PET}, and LPIPS~\cite{LPIPs}, while its
SSIM~\cite{ssim} is nearly identical to the best result at $\lambda=8$.
Moreover, $\lambda=11$ lies within the geometry-supported interval
$[9.10,12.06]$, providing the best empirical trade-off.
We therefore adopt $\lambda=11$ for the MedicalDPO configuration;
the gradient-geometry derivation and cross-seed stability analysis are
provided in the suppl. material.

\section{Conclusion}
\label{sec:conclusion}
We propose MedicalDPO, a preference-alignment framework for Brownian Bridge-based CT-to-PET synthesis. It builds reliable preference pairs covering spatial, structural, and intensity fidelity, and adapts denoising-error preference optimization to Brownian Bridge models via a trainable policy with a frozen reference. Multi-pair rotation and backbone-preserving regularization help balance preference learning with retention of pretrained generation behavior. Experiments on BBDM and CPDM show consistent gains across perceptual, intensity, structural, and reconstruction metrics, with analyses supporting preference reliability. Ablations confirm backbone-preserving regularization is critical to prevent the degradation seen with preference-only fine-tuning. Overall, the results suggest preference optimization can inject medical quality criteria into diffusion-based image translation. Limitations include 2D slice-based synthesis and mainly image-fidelity evaluation. Future work will extend to native 3D translation and clinically relevant endpoints (e.g., uptake quantification and lesion-level analysis), and study medically informed preference weighting, broader sensitivity analyses of preference construction, and sensitivity-aware fine-tuning.


\subsection*{Acknowledge}
The work was partially supported by the following: National Natural Science Foundation of China under No. 92370119; Research Development Fund in XJTLU under no. RDF-23-02-044.
{
    \small
    \bibliographystyle{ieeenat_fullname}
    \bibliography{main}
}

\end{document}


\setcounter{page}{1}
\maketitlesupplementary
\setcounter{table}{4}
\setcounter{figure}{4}
\appendix
\graphicspath{{figures/}}
\section{Additional Method Details}

\subsection{Overall MedicalDPO Training Procedure}

MedicalDPO follows the preference construction and optimization procedure
described in the main paper. For each CT condition, candidates are scored
along spatial, structural, and intensity fidelity dimensions and filtered
into strict preferred--rejected pairs. The pretrained backbone is then duplicated into a trainable policy and
a frozen reference model following preference-based diffusion
fine-tuning~\cite{Original_dpo,diffusion_dpo}, and each pair is evaluated
under matched timestep and Gaussian noise. Fine-tuning
optimizes
\begin{equation}
\mathcal{L}
=
\mathcal{L}_{\rm DPO}
+
\lambda\mathcal{L}_{\rm reg}.
\end{equation}

\subsection{Gradient-Geometry Analysis of \texorpdfstring{$\lambda$}{lambda}}
\label{sec:supp_gradient_lambda}

We perform the following gradient-geometry analysis on
CPDM~\cite{cpdm}, which is also the backbone used for the $\lambda$
sensitivity study in the main paper. For this CPDM-specific analysis,
$\mathcal{L}_{\rm reg}=\mathcal{L}_{\rm CPDM}$. The analysis is an unpreconditioned first-order diagnostic of the raw objective gradients.

\paragraph{Why raw loss magnitudes are insufficient. }
A direct comparison between the numerical values of the DPO and CPDM
losses can be misleading. At initialization, the trainable policy and
the frozen reference model are identical. Therefore, the policy-reference
preference difference is zero and the DPO logit is also zero, giving
\begin{equation}
    \mathcal{L}_{\rm DPO}
    =
    -\log\sigma(0)
    =
    \log 2
    \approx 0.6931.
\end{equation}
Hence, the initial DPO loss contains a substantial constant baseline and
its magnitude should not be directly compared with the CPDM reconstruction
loss for selecting $\lambda$. Since parameter updates are determined by
gradients rather than raw loss values, we instead analyze the geometry of
the two gradients.

\paragraph{Effective-batch gradients.}
Training uses a physical batch size of one with eight gradient-accumulation
steps, giving an effective batch size of eight. To match the actual
optimizer update, the gradient analysis is therefore performed over groups
of eight samples. For an effective batch, we compute
\begin{equation}
    g_D
    =
    \frac{1}{8}
    \sum_{i=1}^{8}
    \nabla_\theta
    \mathcal{L}_{\rm DPO}^{(i)},
    \qquad
    g_C
    =
    \frac{1}{8}
    \sum_{i=1}^{8}
    \nabla_\theta
    \mathcal{L}_{\rm CPDM}^{(i)}.
\end{equation}
The gradient used by the optimizer for
\begin{equation}
    \mathcal{L}
    =
    \mathcal{L}_{\rm DPO}
    +
    \lambda\mathcal{L}_{\rm CPDM}
\end{equation}
is therefore
\begin{equation}
    g_\lambda
    =
    g_D+\lambda g_C.
\end{equation}

\paragraph{Local descent criterion.}
For a sufficiently small learning rate $\eta>0$, the first-order change
of the DPO objective after the update
$\theta'=\theta-\eta g_\lambda$ is
\begin{equation}
    \mathcal{L}_{\rm DPO}(\theta')
    \approx
    \mathcal{L}_{\rm DPO}(\theta)
    -
    \eta\, g_D^\top g_\lambda.
\end{equation}
Thus, a local decrease in the DPO objective requires
\begin{equation}
    g_D^\top g_\lambda>0.
\end{equation}
Applying the same argument to the CPDM objective gives
\begin{equation}
    g_C^\top g_\lambda>0.
\end{equation}
We therefore seek values of $\lambda$ that satisfy both conditions,
so that one optimizer update is locally compatible with both preference
learning and preservation of the pretrained CPDM objective.

Let
\begin{equation}
    A=\|g_D\|_2^2,
    \qquad
    B=\|g_C\|_2^2,
    \qquad
    C=g_D^\top g_C.
\end{equation}
The two descent conditions become
\begin{equation}
    A+\lambda C>0,
    \qquad
    C+\lambda B>0.
\end{equation}
When $C<0$, the gradients are locally conflicting and these inequalities
impose both a lower and an upper constraint on $\lambda$:
\begin{equation}
    -\frac{C}{B}
    <
    \lambda
    <
    -\frac{A}{C}.
\end{equation}
The lower bound prevents insufficient CPDM regularization, whereas the
upper bound prevents the regularization gradient from dominating the DPO
descent direction.

\paragraph{Pilot and stability analysis.}
We first conduct a pilot experiment using 64 randomly sampled training
examples, corresponding to eight effective batches. Five of the eight
batches exhibit gradient conflict. The intersection of their local descent
constraints gives
\begin{equation}
    5.1455 < \lambda < 39.5556.
\end{equation}
Among the tested candidate values, $\lambda=6$ is the smallest value for
which all eight effective batches satisfy both descent conditions.
However, eight effective batches provide limited evidence and are
sensitive to the sampled examples. We therefore perform a larger stability
analysis using 256 samples for each of three random seeds
(42, 1234, and 2025), yielding 32 effective batches per seed.

\begin{table}[t]
\centering
\caption{
Gradient-geometry stability across random subsets.
Each effective batch contains eight samples, matching the
gradient-accumulation setting used during training.
The strict interval denotes the range satisfying both local descent
conditions for all conflicting batches within the corresponding run.
}
\label{tab:supp_gradient_stability}
\small
\setlength{\tabcolsep}{4pt}
\begin{tabular}{ccccc}
\toprule
Seed & Samples & Batches & Conflict & Strict interval \\
\midrule
42 (pilot) & 64  & 8  & 5  & $(5.15,\,39.56)$ \\
42         & 256 & 32 & 14 & $(5.15,\,14.38)$ \\
1234       & 256 & 32 & 22 & $(9.10,\,12.06)$ \\
2025       & 256 & 32 & 16 & $(13.25,\,23.99)$ \\
\bottomrule
\end{tabular}
\end{table}
As summarized in Tab.~\ref{tab:supp_gradient_stability}, the strict
feasible interval varies substantially across random subsets. The three
256-sample runs do not share a common strict intersection, indicating
that no single $\lambda$ satisfies every local descent constraint across
all sampled effective batches. We therefore treat gradient geometry as a
screening criterion rather than as a universal guarantee for
hyperparameter selection.

\begin{figure}[t]
    \centering
    \includegraphics[width=\linewidth]{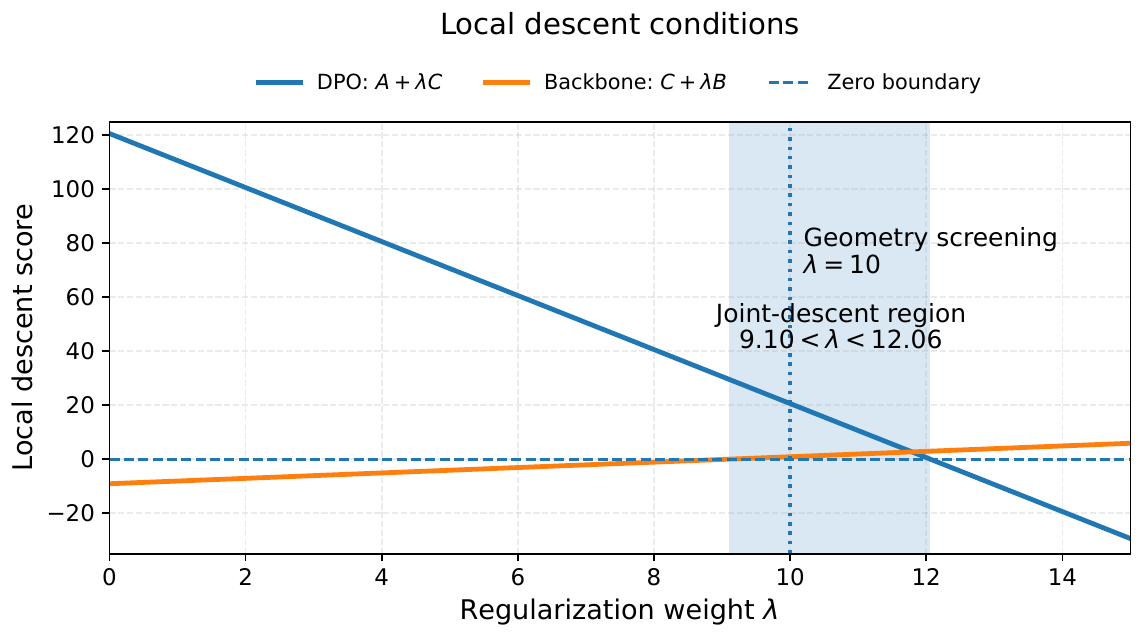}
    \caption{
Illustrative instantiation of the local descent conditions using the
seed-1234 gradient-geometry bounds. The shaded region denotes the strict
feasible interval $9.10<\lambda<12.06$, within which both the DPO and
CPDM objectives satisfy the local descent conditions for the corresponding
run. The vertical marker highlights $\lambda=10$ as the
geometry-screened candidate. The vertical scale is illustrative; the zero crossings and the resulting
feasible interval are the quantities of interest.
}
    \label{fig:supp_lambda_local}
\end{figure}

The interval $9.10<\lambda<12.06$ reported in the main paper corresponds
to the 256-sample run with seed 1234, which yields the narrowest strict
interval among the three stability runs. For visualization, Fig.~\ref{fig:supp_lambda_local} provides an
illustrative instantiation of the local descent geometry using these
seed-1234 bounds.

\paragraph{Cross-seed descent coverage.}
To identify a value that behaves consistently across random subsets, we
pool the three 256-sample analyses, yielding 96 effective batches, and
measure the fraction for which both objectives satisfy the local descent
criterion. We additionally track DPO descent separately to detect cases
where excessive regularization begins to suppress the preference update.

\begin{table}[t]
\centering
\caption{
Cross-seed gradient-descent coverage over 96 effective batches from
three random seeds. Joint descent indicates that both DPO and CPDM
objectives satisfy their local descent conditions. Here, $\lambda=10$ denotes the geometry-screened candidate; the final
MedicalDPO configuration uses $\lambda=11$ based on the empirical
sensitivity analysis reported in the main paper.
}
\label{tab:supp_lambda_coverage}
\small
\setlength{\tabcolsep}{7pt}
\begin{tabular}{ccc}
\toprule
$\lambda$ & Joint descent & DPO descent \\
\midrule
6    & 92/96 (95.8\%) & 96/96 (100\%) \\
10   & 95/96 (99.0\%) & 96/96 (100\%) \\
13   & 94/96 (97.9\%) & 95/96 (99.0\%) \\
13.5 & 95/96 (99.0\%) & 95/96 (99.0\%) \\
14   & 95/96 (99.0\%) & 95/96 (99.0\%) \\
\bottomrule
\end{tabular}
\end{table}

\begin{figure}[t]
    \centering
    \includegraphics[width=\linewidth]{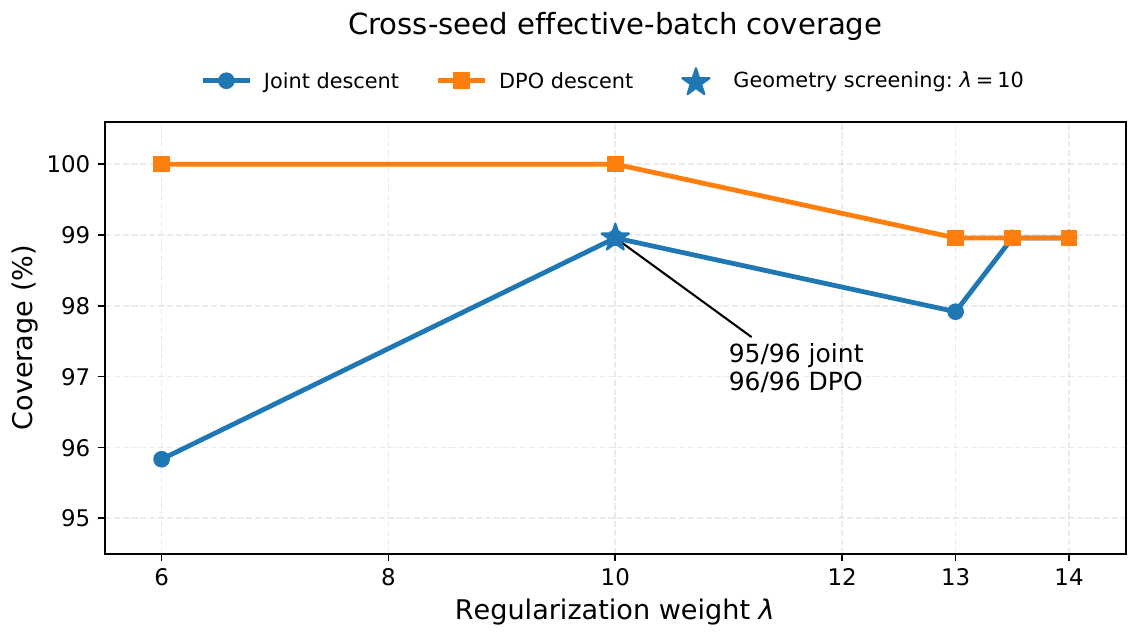}
    \caption{
    Cross-seed effective-batch coverage over 96 effective batches from
    three 256-sample runs. Joint descent requires both the DPO and CPDM
    objectives to satisfy their local descent conditions, while DPO
    descent considers the preference objective alone.
    The geometry-screened candidate $\lambda=10$ attains the maximum
observed joint-descent coverage (95/96) while preserving DPO descent
for all 96 effective batches.
    }
    \label{fig:supp_lambda_coverage}
\end{figure}

As shown in Tab.~\ref{tab:supp_lambda_coverage} and
Fig.~\ref{fig:supp_lambda_coverage}, $\lambda=10$ achieves the maximum
observed joint-descent coverage of 95/96 effective batches while
preserving DPO descent for all 96 batches. Larger values such as
$\lambda=13.5$ and $\lambda=14$ reach the same joint coverage but cause
one effective batch to violate the DPO descent condition. Thus,
$\lambda=10$ provides the most balanced choice according to the
gradient-geometry criterion alone.

Importantly, the gradient-geometry analysis is used only to identify a
plausible regularization regime rather than to determine the final
hyperparameter solely from local gradients. The final MedicalDPO
configuration uses $\lambda=11$, following the empirical sensitivity
analysis reported in the main paper.

\subsection{Detailed Formulations of Ablation Variants}

We provide the detailed objectives corresponding to the ablation
variants reported in the main paper.

\paragraph{DPO.}
This variant implements a Brownian-Bridge adaptation of the
Diffusion-DPO-style relative denoising preference objective
~\cite{diffusion_dpo}
using the same preference pairs as MedicalDPO but without the proposed
backbone-preserving regularization:
\begin{equation}
    \mathcal{L}
    =
    \mathcal{L}_{\rm DPO}.
\end{equation}

\paragraph{Only GT.}
This control performs additional fine-tuning using only the original
ground-truth training objective:
\begin{equation}
    \mathcal{L}
    =
    \mathcal{L}_{\rm GT}.
\end{equation}
Here, $\mathcal{L}_{\rm GT}$ denotes the original training loss of the
corresponding Brownian Bridge backbone evaluated using the ground-truth
PET target. 

\paragraph{DPO + Guard.}
This variant penalizes degradation of the preferred sample relative to the
frozen reference model:
\begin{equation}
    \mathcal{L}
    =
    \mathcal{L}_{\rm DPO}
    +
    \lambda_{\rm guard}\mathcal{L}_{\rm guard},
\end{equation}
where
\begin{equation}
    \mathcal{L}_{\rm guard}
    =
    \operatorname{ReLU}
    \left(
        \ell_{\rm BB,\theta}(y_w)
        -
        \ell_{\rm BB,ref}(y_w)
    \right),
\end{equation}
with $\lambda_{\rm guard}=11$. For brevity, the conditioning input,
timestep, and noise variables of $\ell_{\rm BB}$ are omitted here.

\paragraph{DPO + GT + Drift.}
This variant combines preference optimization with ground-truth
supervision and an explicit policy-reference drift penalty:
\begin{equation}
    \mathcal{L}
    =
    \mathcal{L}_{\rm DPO}
    +
    11\mathcal{L}_{\rm GT}
    +
    1.1\mathcal{L}_{\rm drift},
\end{equation}
where
\begin{equation}
    \mathcal{L}_{\rm drift}
    =
    \frac{1}{d}
    \left\|
        \hat g_\theta^{\rm gt}
        -
        \hat g_{\rm ref}^{\rm gt}
    \right\|_2^2.
\end{equation}
Here, $\hat g_\theta^{\rm gt}$ and $\hat g_{\rm ref}^{\rm gt}$ denote the
denoising predictions of the trainable and reference models under
ground-truth supervision, respectively.

\paragraph{Winner-to-GT.}
This strategy additionally constrains the preferred sample toward the
ground-truth PET:
\begin{equation}
    \mathcal{L}
    =
    \mathcal{L}_{\rm DPO}
    +
    \lambda_{\rm gt}
    \mathcal{L}_{w\rightarrow{\rm GT}},
\end{equation}
where
\begin{equation}
    \mathcal{L}_{w\rightarrow{\rm GT}}
    =
    \left\|y_w-y_{\rm gt}\right\|_1,
\end{equation}
and $\lambda_{\rm gt}=11$.

\paragraph{Winner/Loser-to-GT.}
We further examine objectives that apply ground-truth constraints to both
members of the preference pair:
\begin{equation}
    \mathcal{L}
    =
    \mathcal{L}_{\rm DPO}
    +
    \lambda_{\rm gt}\mathcal{L}_{w\rightarrow{\rm GT}}
    +
    \lambda_{\rm losegt}\mathcal{L}_{l\rightarrow{\rm GT}}.
\end{equation}
Here,
\begin{equation}
    \mathcal{L}_{l\rightarrow{\rm GT}}
    =
    \left\|y_l-y_{\rm gt}\right\|_1,
\end{equation}
while $\mathcal{L}_{w\rightarrow{\rm GT}}$ is defined above.
Two weighting configurations are evaluated:
\begin{equation}
    \mathrm{(A)}:
    (\lambda_{\rm gt},\lambda_{\rm losegt})=(11,11),
\end{equation}
and
\begin{equation}
    \mathrm{(B)}:
    (\lambda_{\rm gt},\lambda_{\rm losegt})=(11,1.1).
\end{equation}

\section{Additional Experimental Details}
\subsection{Additional Preference Split Statistics}
\label{sec:supp_datasets}

After strict preference filtering, the retained preference conditions are
split into training and validation sets for each backbone, as summarized
in Tab.~\ref{tab:supp_preference_statistics}.

\FloatBarrier
\begin{table}[t]
\centering
\caption{Additional statistics of the retained 15K preference sets.}
\label{tab:supp_preference_statistics}
\small
\begin{tabular}{lccc}
\toprule
Backbone & Train CTs & Val. CTs & Retained CTs \\
\midrule
CPDM & 12,704 & 1,412 & 14,116 \\
BBDM & 12,685 & 1,409 & 14,094 \\
\bottomrule
\end{tabular}
\end{table}

\subsection{Backbone-Specific Preprocessing}
\label{sec:supp_preprocessing}

For BBDM~\cite{bbdm}, CT and PET intensities are clipped to
modality-specific maximum values of 2,047 and 32,767, respectively,
divided by the corresponding maximum, and subsequently mapped to
$[-1,1]$. Area interpolation is used for downsampling and bilinear
interpolation for upsampling.

For CPDM~\cite{cpdm}, CT intensities are divided by 2,047 without the
subsequent $[-1,1]$ mapping, whereas PET intensities are divided by
32,767 and further mapped to $[-1,1]$.

\subsection{Attention Map Implementation}
\label{sec:supp_attention}

Beyond the CPDM attention-map generation pipeline~\cite{cpdm} described
in the main paper, the segmentation logits are passed through a sigmoid
function, and the resulting continuous probability maps are saved directly
without binarization or additional min--max normalization.
These attention maps are precomputed before MedicalDPO fine-tuning
and remain fixed throughout preference optimization. For each CT
condition, the same precomputed attention map is used to score all
generated PET candidates.

During preference scoring, the saved attention maps are loaded and used
directly without further normalization, clipping, or resizing; their
spatial dimensions are required to match those of the corresponding PET
images. Attention values outside $[0,1]$ are explicitly monitored.
If the sum of the attention weights is no greater than $10^{-8}$, the
weighted error is marked as invalid rather than performing division by
zero.

\subsection{Base Backbone Training Details}
\label{sec:supp_backbones}

Additional training settings for the pretrained BBDM~\cite{bbdm} and
CPDM~\cite{cpdm} backbones are summarized in
Tab.~\ref{tab:supp_backbone_settings}.

\begin{table}[t]
\centering
\caption{Additional training configurations of the diffusion backbones.}
\label{tab:supp_backbone_settings}
\small
\setlength{\tabcolsep}{5pt}
\begin{tabular}{lcc}
\toprule
Setting & BBDM & CPDM \\
\midrule
Training epochs & 100 & 200 \\
Batch size & 4 & 16 \\
Gradient accumulation & 2 & 4 \\
Optimizer & Adam & Adam \\
Learning rate & $1\times10^{-4}$ & $1\times10^{-4}$ \\
EMA decay & 0.995 & 0.995 \\
Selected checkpoint & Epoch 90 & Epoch 136 \\
\bottomrule
\end{tabular}
\end{table}

\subsection{Additional MedicalDPO Training Details}
\label{sec:supp_training}

For both BBDM and CPDM, MedicalDPO updates only the denoising network,
while the remaining backbone components are frozen. Weight decay is
$10^{-2}$ and the gradient norm is clipped to 1.0. BBDM and CPDM use
linear warmup ratios of 0.10 and 0.05, respectively, followed by a
constant learning rate.

All hyperparameters and checkpoints are selected using validation data. For the $\lambda$-sensitivity study in Fig.
~4 of the main paper, the 2.5K preference subset is sampled exclusively from the training pool, while model comparison and hyperparameter selection are performed on the separate validation set. The held-out test set of 246 paired CT-PET images is excluded from this process and used only for final evaluation.

\subsection{Evaluation Implementation Details}
\label{sec:supp_metrics}

All evaluation metrics are computed on PET images normalized by the fixed
maximum intensity of 32,767. SSIM~\cite{ssim} and
PSNR~\cite{medgan,uncertainty_guidedgan} are computed with
\texttt{data\_range}=1. MAE~\cite{Synthetic_PET} is first evaluated on the
normalized images and then multiplied by 32,767 to report the error on
the original PET intensity scale.

LPIPS~\cite{LPIPs} is computed using the VGG backbone. The normalized
grayscale PET images are mapped from $[0,1]$ to $[-1,1]$ and replicated
across three channels before being passed to the LPIPS network. Metrics
are computed per image and averaged within each seed. Standard deviations
are computed across the three seed-level averages.

\begin{table}[!tb]
\centering
\caption{
Sensitivity to the DPO scaling parameter $\beta$ for preference-only Rotation fine-tuning on CPDM at epoch 20. Within the tested range, increasing $\beta$ progressively reduces the degradation caused by DPO-only fine-tuning, but no setting recovers the pretrained Base performance across all four metrics.  Best DPO values are shown in
bold.
}
\label{tab:supp_beta}
\small
\setlength{\tabcolsep}{3.5pt}
\begin{tabular}{lcccc}
\toprule
Setting
& SSIM $\uparrow$
& PSNR $\uparrow$
& MAE $\downarrow$
& LPIPS $\downarrow$ \\
\midrule
Base
& 0.84994
& 24.26565
& 751.05
& 0.09758 \\

$\beta=50$
& 0.80122
& 19.79904
& 1424.11
& 0.13410 \\

$\beta=100$
& 0.81351
& 20.45311
& 1298.82
& 0.12433 \\

$\beta=200$
& 0.82530
& 21.44949
& 1126.31
& 0.11686 \\

$\beta=400$
& 0.83479
& 22.74217
& 932.35
& 0.11176 \\

$\beta=800$
& \textbf{0.84330}
& \textbf{23.84394}
& \textbf{788.02}
& \textbf{0.10822} \\
\bottomrule
\end{tabular}
\end{table}

\begin{table*}[!tb]
\centering
\caption{
Seed-wise relative improvements over the corresponding Base model.
Positive values denote improvements; Mean $\pm$ Std is computed across
the three random seeds.
}
\label{tab:supp_seed_results}

\footnotesize
\setlength{\tabcolsep}{6pt}
\renewcommand{\arraystretch}{1.10}

\begin{tabular}{llccccc}
\toprule
Backbone & Strategy & Seed
& LPIPS
& MAE
& SSIM
& PSNR \\
\midrule

\multirow{8}{*}{BBDM}
& Single & 42
& $+11.41\%$
& $+12.31\%$
& $+2.24\%$
& $+3.02\%$ \\

& & 1234
& $+8.31\%$
& $+9.94\%$
& $+1.76\%$
& $+2.25\%$ \\

& & 2025
& $+13.28\%$
& $+13.53\%$
& $+2.64\%$
& $+3.11\%$ \\

& & Mean $\pm$ Std
& $10.998 \pm 2.512\%$
& $11.928 \pm 1.827\%$
& $2.214 \pm 0.438\%$
& $2.789 \pm 0.472\%$ \\

\cmidrule(lr){2-7}

& Rotation & 42
& $+12.30\%$
& $+13.82\%$
& $+2.30\%$
& $+3.53\%$ \\

& & 1234
& $+8.79\%$
& $+11.10\%$
& $+1.77\%$
& $+2.69\%$ \\

& & 2025
& $+13.85\%$
& $+14.71\%$
& $+2.68\%$
& $+3.57\%$ \\

& & Mean $\pm$ Std
& $11.646 \pm 2.594\%$
& $13.208 \pm 1.879\%$
& $2.248 \pm 0.457\%$
& $3.262 \pm 0.494\%$ \\

\midrule

\multirow{8}{*}{CPDM}
& Single & 42
& $+0.97\%$
& $+5.34\%$
& $+0.58\%$
& $+0.71\%$ \\

& & 1234
& $+0.62\%$
& $+4.93\%$
& $+0.55\%$
& $+0.53\%$ \\

& & 2025
& $+0.97\%$
& $+5.43\%$
& $+0.59\%$
& $+0.68\%$ \\

& & Mean $\pm$ Std
& $0.856 \pm 0.202\%$
& $5.233 \pm 0.265\%$
& $0.573 \pm 0.025\%$
& $0.640 \pm 0.096\%$ \\

\cmidrule(lr){2-7}

& Rotation & 42
& $+0.80\%$
& $+3.28\%$
& $+0.45\%$
& $+0.20\%$ \\

& & 1234
& $+0.43\%$
& $+2.93\%$
& $+0.42\%$
& $+0.04\%$ \\

& & 2025
& $+0.79\%$
& $+3.38\%$
& $+0.46\%$
& $+0.17\%$ \\

& & Mean $\pm$ Std
& $0.677 \pm 0.212\%$
& $3.198 \pm 0.238\%$
& $0.443 \pm 0.022\%$
& $0.137 \pm 0.085\%$ \\

\bottomrule
\end{tabular}
\end{table*}
\section{Additional Experimental Results}
\subsection{Sensitivity to \texorpdfstring{$\beta$}{beta}}
\label{sec:supp_beta}

We examine whether the degradation caused by preference-only fine-tuning can be mitigated by adjusting the DPO scaling parameter $\beta$~\cite{Original_dpo,diffusion_dpo}. On the CPDM backbone, we evaluate
$\beta\in\{50,100,200,400,800\}$ using the Rotation strategy without the backbone-preserving objective. All variants are compared at epoch 20 under the same training configuration.

As shown in Tab.~\ref{tab:supp_beta}, increasing $\beta$ progressively improves all four
metrics within the evaluated range. SSIM and PSNR increase monotonically, while MAE and LPIPS decrease. Nevertheless,  even $\beta=800$, the strongest DPO-only setting, remains inferior to the pretrained CPDM backbone across all four metrics. 

These results indicate that tuning $\beta$ can reduce the
degradation introduced by preference-only DPO, but does not fully
preserve the pretrained synthesis fidelity. This observation motivates
explicitly retaining the original backbone objective during preference
fine-tuning, as done in MedicalDPO through
$\mathcal{L}_{\rm DPO}+\lambda\mathcal{L}_{\rm reg}$.
Accordingly, this ablation is used as diagnostic evidence rather than
to select a large $\beta$ for the final model; the final MedicalDPO
experiments retain $\beta=100$ and instead regulate preference
adaptation through the backbone-preserving term.

\subsection{Additional Preference Analysis}
\label{sec:supp_preference}

To further examine the reliability of the proposed preference
construction, we provide an additional quantitative analysis on
BBDM-generated candidates~\cite{bbdm}. Table~\ref{tab:supp_bbdm_preference}
summarizes the quality separation between the retained preferred and
rejected candidates after strict multi-objective filtering.

\begin{table}[t]
\centering
\caption{
Preference-quality analysis on BBDM-generated candidates.
Preferred and rejected values are reported as mean $\pm$ standard
deviation, and $\Delta$ denotes Preferred minus Rejected.
MAE and WMAE are computed on normalized PET intensities.
}
\label{tab:supp_bbdm_preference}
\small
\setlength{\tabcolsep}{4pt}
\begin{tabular}{lccc}
\toprule
Metric & Preferred & Rejected & $\Delta$ \\
\midrule
Dice
& $0.4575\pm0.2355$
& $0.1318\pm0.1759$
& $+0.3258$ \\

SSIM
& $0.9053\pm0.0414$
& $0.8566\pm0.1135$
& $+0.0487$ \\

MAE
& $0.0107\pm0.0053$
& $0.0204\pm0.0114$
& $-0.0096$ \\

WMAE
& $0.0349\pm0.0287$
& $0.0710\pm0.0639$
& $-0.0361$ \\
\bottomrule
\end{tabular}
\end{table}

As shown in Tab.~\ref{tab:supp_bbdm_preference}, the retained BBDM
preference pairs exhibit clear and consistent quality separation.
Preferred candidates achieve substantially higher Dice and SSIM than
rejected candidates, with improvements of $0.3258$ and $0.0487$,
respectively. They also reduce MAE and WMAE by $0.0096$ and $0.0361$.
The consistent improvement across spatial, structural, and intensity
fidelity indicates that the proposed preference construction remains
effective when applied to BBDM-generated candidates.

\begin{figure*}[t]
    \centering
    \includegraphics[width=\textwidth]{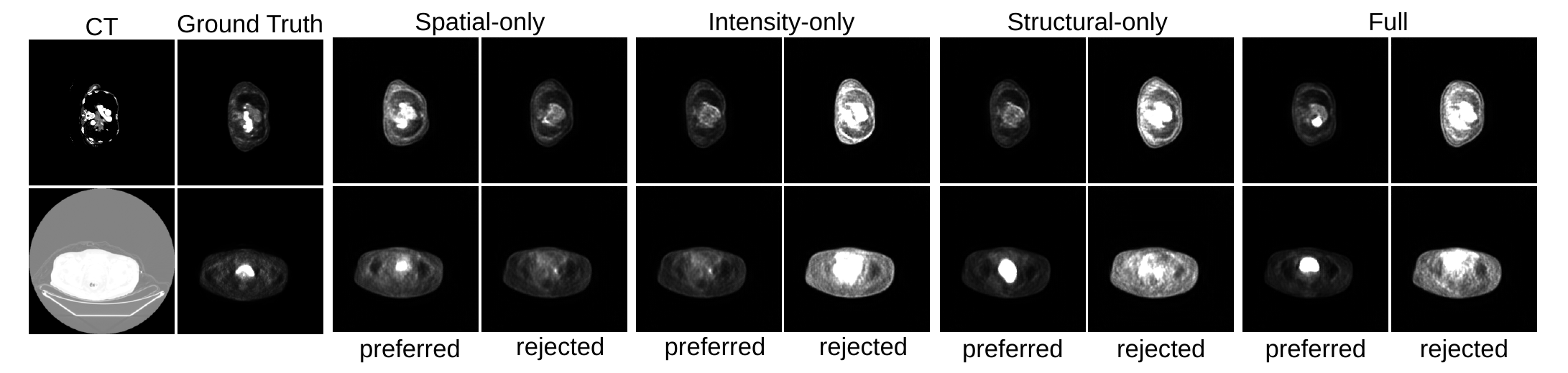}
    \caption{
    Additional qualitative examples of preference construction on
    CPDM-generated candidates. Each row shows one CT condition, its
    ground-truth PET, and the preferred/rejected candidates selected by
    spatial-only, intensity-only, structural-only, and full
    multi-objective strategies. Single-objective criteria can select
    different preference pairs because they emphasize complementary
    aspects of PET fidelity, whereas the full strategy combines all
    three fidelity dimensions and applies strict cross-metric
    verification.
    }
    \label{fig:supp_preference_vis}
\end{figure*}

We further provide two qualitative examples from CPDM-generated
candidate pools~\cite{cpdm} in Fig.~\ref{fig:supp_preference_vis}. For the same CT
condition, spatial-only, intensity-only, and structural-only criteria
can select noticeably different preferred and rejected candidates,
reflecting the complementary fidelity aspects emphasized by individual
objectives. In contrast, the full strategy jointly considers spatial,
intensity, and structural fidelity and applies strict cross-metric
verification to construct cross-metric-consistent preference pairs.

Together, the quantitative BBDM analysis and qualitative CPDM examples
provide additional evidence that the proposed multi-objective
preference construction yields reliable preferred--rejected pairs
across different Brownian Bridge generation backbones.

\subsection{Robustness Across Random Seeds}
To further assess the robustness of MedicalDPO to stochastic sampling,
we report the seed-wise relative improvements over the corresponding
pretrained backbones in Tab.~\ref{tab:supp_seed_results}. The main paper
reports results averaged over three random seeds, whereas here we
separately analyze seeds 42, 1234, and 2025. For LPIPS~\cite{LPIPs} and MAE~\cite{Synthetic_PET}, the
relative improvement is computed as the percentage reduction from the
corresponding Base result; for SSIM~\cite{ssim} and PSNR~\cite{medgan,uncertainty_guidedgan}, it is computed as the
percentage increase. Therefore, positive values indicate improvement for
all four metrics.

As shown in Tab.~\ref{tab:supp_seed_results}, MedicalDPO consistently
improves the corresponding Base model across all three random seeds.
For BBDM~\cite{bbdm}, both Single and Rotation yield substantial improvements in all
four metrics, indicating that the gains reported in the main paper are
not driven by a particular stochastic sample. Rotation generally provides
the larger improvement, especially in LPIPS, MAE, and PSNR.

The same trend is observed for CPDM~\cite{cpdm}, although the magnitude of improvement
is smaller. Importantly, both Single and Rotation improve all four metrics
for every evaluated seed. In particular, Single consistently reduces MAE
by approximately $5\%$ while also improving LPIPS, SSIM, and PSNR. These
results show that the improvements on CPDM remain systematic across
different stochastic sampling conditions rather than arising from a
single favorable seed.

\subsection{Additional Qualitative Results}
\label{sec:supp_qualitative}

\begin{figure*}[!t]
    \centering
    \includegraphics[width=0.92\textwidth]{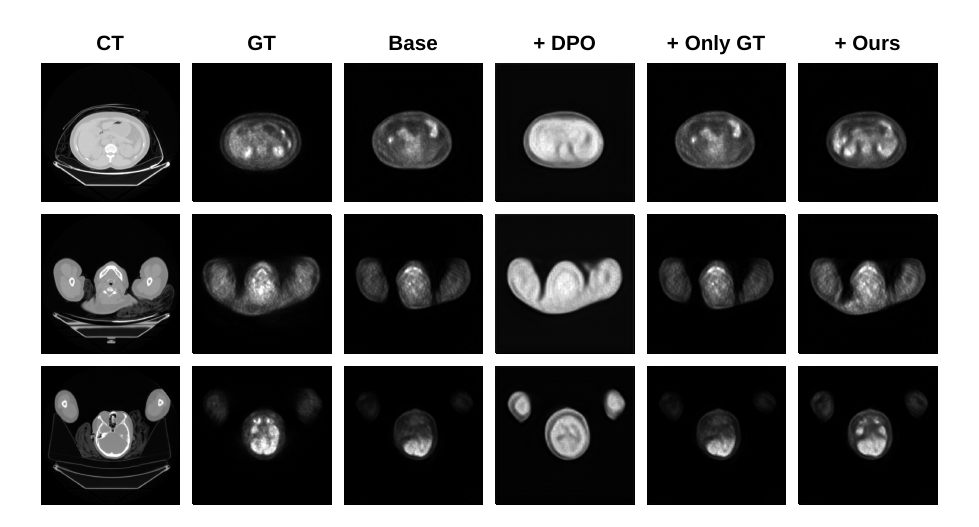}
   \caption{Additional qualitative comparisons on BBDM~\cite{bbdm}.
$+$Ours denotes MedicalDPO with the Rotation strategy.
}
    \label{fig:supp_bbdm}
\end{figure*}

\begin{figure*}[!t]
    \centering
    \includegraphics[width=0.92\textwidth]{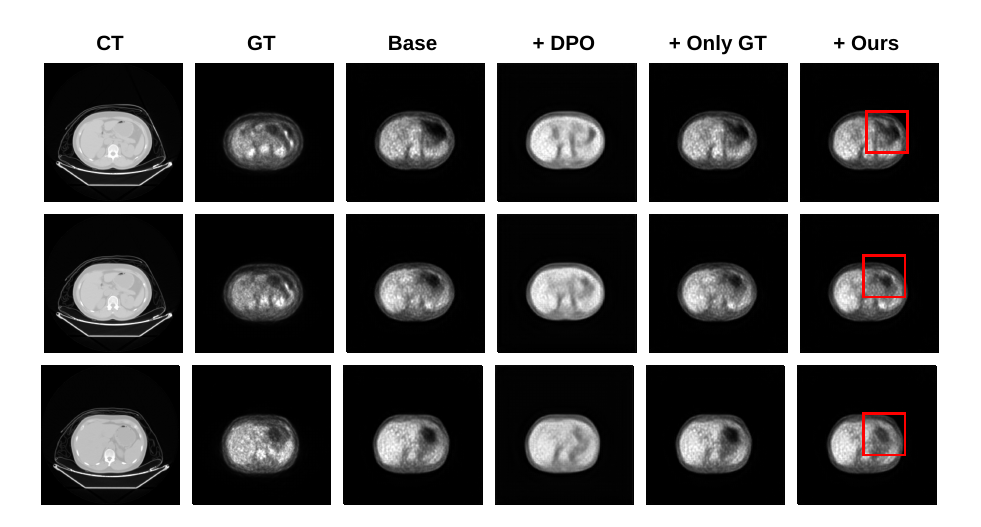}
    \caption{
Additional qualitative comparisons on CPDM~\cite{cpdm}.
$+$Ours denotes MedicalDPO with the Single strategy.
Red boxes highlight regions where MedicalDPO improves local uptake details.
}
    \label{fig:supp_cpdm}
\end{figure*}

We provide additional qualitative comparisons on BBDM~\cite{bbdm} and
CPDM~\cite{cpdm} in Figs.~\ref{fig:supp_bbdm} and
\ref{fig:supp_cpdm}, respectively. Each example compares the input CT,
ground-truth PET, the pretrained backbone, preference optimization without
the proposed regularization, additional ground-truth fine-tuning, and the
final MedicalDPO result.

As shown in Fig.~\ref{fig:supp_bbdm}, direct preference optimization
without the proposed regularization can substantially alter the pretrained
BBDM generation behavior, often producing overly bright or spatially
distorted uptake patterns. In contrast, $+$Only GT remains visually close
to the base model and introduces only limited changes. In these examples, $+$Ours produces PET images with closer visual agreement to the paired ground-truth PET in overall structure and intensity distribution. These examples illustrate the visual changes introduced by MedicalDPO and complement the quantitative BBDM results.

Fig.~\ref{fig:supp_cpdm} shows the corresponding comparison on
CPDM~\cite{cpdm}. Since the pretrained CPDM already provides relatively
strong overall synthesis quality, the improvements introduced by
MedicalDPO are mainly reflected in finer local details. As highlighted by
the red boxes, $+$Ours produces local intensity patterns with closer visual agreement to the paired ground-truth PET while largely preserving the overall structure of the Base prediction. In comparison,
$+$DPO can disturb the local uptake distribution, whereas $+$Only GT
produces only limited changes. These results suggest that MedicalDPO
further refines local PET details on top of the strong CPDM backbone.

{
    \small
    \bibliographystyle{ieeenat_fullname}
    \bibliography{main}
}